\pdfoutput=1

\documentclass[11pt]{article}

\usepackage[]{EMNLP2022}

\usepackage{times}
\usepackage{latexsym}
\usepackage{graphicx}
\usepackage{multirow}
\usepackage{float}
\usepackage{subfig}
\usepackage{amsmath}
\usepackage{algorithm}
\usepackage{amssymb}
\usepackage{color}
\usepackage{makecell}
\usepackage{algorithm}
\usepackage{algorithmicx}
\usepackage{algpseudocode}
\usepackage{blindtext}
\usepackage{paralist}
\usepackage{enumitem}
\usepackage{booktabs} 
\setlist{topsep=0pt, leftmargin=*}

\usepackage[T1]{fontenc}

\usepackage[utf8]{inputenc}

\usepackage{microtype}

\usepackage{bm}

\def\bW{\textbf{W}}
\def\bal{\bm{\alpha}}
\DeclareMathOperator*{\argmin}{arg\,min}
\DeclareMathOperator*{\argmax}{arg\,max}

\title{Search to Pass Messages for Temporal Knowledge Graph Completion}

\author{
Zhen Wang$^{1,2}$ \quad  
Haotong Du$^{1,2,}$\footnotemark[1] \quad 
Quanming Yao$^{3,}$\footnotemark[1] \quad 
Xuelong Li$^2$ \\
	$^1$School of Computer Science, Northwestern Polytechnical University, China \\
	$^2$School of Artificial Intelligence, Optics and Electronics (iOPEN), \\ Northwestern Polytechnical University, China \\
	$^3$Department of Electronic Engineering, Tsinghua University, China \\
	\texttt{w-zhen@nwpu.edu.cn,}
	\texttt{duhaotong@mail.nwpu.edu.cn}\\
	\texttt{qyaoaa@tsinghua.edu.cn,}
	\texttt{li@nwpu.edu.cn}}

\begin{document}
\maketitle{
\renewcommand{\thefootnote}{\fnsymbol{footnote}}
\setcounter{footnote}{0}
\footnotetext[1]{Corresponding author.}
}
\begin{abstract}
	
Completing missing facts is a fundamental task for temporal knowledge graphs (TKGs).
Recently,
graph neural network (GNN) based methods, 
which can simultaneously explore topological and temporal information, 
have become the state-of-the-art (SOTA) to complete TKGs. 
However, 
these studies are based on hand-designed architectures and fail to explore the diverse topological and temporal properties of TKG.
To address this issue, 
we propose to use neural architecture search (NAS) to design data-specific message passing architecture for TKG completion.
In particular, 
we develop a generalized framework to explore topological and temporal information in TKGs.
Based on this framework, we design an expressive search space to fully capture various properties of different TKGs. 
Meanwhile,
we adopt a search algorithm, which trains a supernet structure by sampling single path for efficient search with less cost.
We further conduct extensive experiments on three benchmark datasets.
The results show that the searched architectures by our method achieve the SOTA performances.
Besides, 
the searched models can also implicitly reveal diverse properties in different TKGs.
Our code is released in \url{https://github.com/striderdu/SPA}. 

\end{abstract}

\section{Introduction}

A temporal knowledge graph (TKG)~\citep{cai2022temporal}
is a graph-structural data with many time-sensitive relational facts. 
The facts can be formed as qudaruples 
\textit{(subject entity, relationship, object entity, timestamp)}, 
denoted as $(s, r, o, t)$, 
e.g., 
\emph{(FIFA World Cup, is held in, Qatar, 2022)}. 
TKGs are used extensively in various applications that 
require the assistance of temporal knowledge 
such as 
temporal question answering~\citep{saxena-etal-2021-question}, 
recommendation systems~\citep{zhao2021time} 
and mobility prediction~\citep{wang2021spatio}.

Notably, 
similar to static KG, 
most TKGs are inherently incompletion, 
which seriously hampers their applications in downstream tasks.
Therefore, 
a great number of works focus on TKG completion (TKGC) to infer the missing facts in TKGs.
Pioneer embedding-based methods~\citep{leblay2018deriving,dasgupta-etal-2018-hyte,goel2020diachronic,lacroix2020tensor} 
directly construct time-aware score functions 
to evaluate the plausibility of quadruple.
However, 
embedding-based methods 
do not explicitly encode local graph structures in TKG, 
which limits their expressiveness.

Recently,
based on the success of graph neural networks (GNNs), 
some GNN-based methods have been proposed to solve TKGC.
TeMP~\citep{wu-etal-2020-temp}, 
a typical GNN-based method for TKGC, 
discretizes a TKG into multiple static KG snapshots and generates dynamic entity representations along two dimensions: structural neighborhoods and temporal dynamics. 
Structural encoder extracts feature from local node neighborhoods in each snapshot through message passing and aggregation, 
while temporal encoder captures feature evolution over multiple time steps by sequential models. 
T-GAP~\citep{jung2021learning}, 
views timestamps as properties of links between entities, 
and proposes the temporal GNN to learn structural and temporal information on the whole graph.  
GNNs have been demonstrated to achieve better performance for TKGC tasks, 
due to their powerful expressiveness.

However, 
these GNN-based methods 
use the fixed GNN architectures to tackle different TKGs, 
failing to explore the diverse topological and temporal properties of TKGs, 
which prevents the model from fully discovering the diverse implicit patterns in different datasets. 
More recently, 
BoxTE~\citep{messner2022temporal} has also pointed out this problem. 
Therefore, 
it is critical to design data-specific GNN architectures for TKGC task.

Neural architecture search (NAS)~\citep{yao2018taking, hutter2019automated} 
has achieved great success in designing data-specific architectures, 
of which the performances exceed the architectures crafted by human experts in various areas, 
e.g., computer vision~\citep{zhang2022searching}, 
natural language processing~\citep{so2019evolved}, 
and graph learning~\citep{zhang2021automated}. 
More recently, 
in static KG completion, 
there are some works that adopt NAS techniques for designing the score function~\citep{zhang2022bilinear} 
or GNN architecture~\citep{wang2021autogel}. 
However, 
no one has made similar attempts on TKG. 
And designing data-specific architectures for TKGC task is non-trivial, 
because of the demand to simultaneously explore topological and temporal information.

In this work, 
we propose a novel method which tries to Search to PAss messages(SPA), 
to automatically design data-specific architectures for TKGC. 
Firstly, 
we design a generalized framework 
to simultaneously explore topological and temporal information in TKGs. 
From this, 
we define a novel and expressive search space, 
in which different combinations of operations can capture various patterns of different TKGs. 
To enable efficient search on top of the search space, 
we adopt a flexible and effective search algorithm, 
which trains a supernet by sampling single path uniformly, 
thus greatly reducing the GPU memory cost. 
To demonstrate the effectiveness of SPA, 
we conduct extensive experiments on three benchmark datasets of TKGC. 
Experimental results show that SPA can consistently achieve state-of-the-art performance by designing data-specific architectures. 
Further empirical results verify the searched models provide implicitly properties expression for different TKGs.

\section{The Proposed Method}
As mentioned in the introduction, 
the GNN-based method for TKGC should be data-specific.
Generally, 
TKGs contain both topological and temporal information. 
Thus, 
to design a proper model, 
we first define a framework which can model topological patterns and temporal contexts jointly. 
Then, 
we introduce our novel search space.
Finally, we describe our search objective and search algorithm.

\subsection{The Generalized Framework}\label{sec-framework}

\begin{figure}[t]
	\centering
	\includegraphics[width=0.9\linewidth]{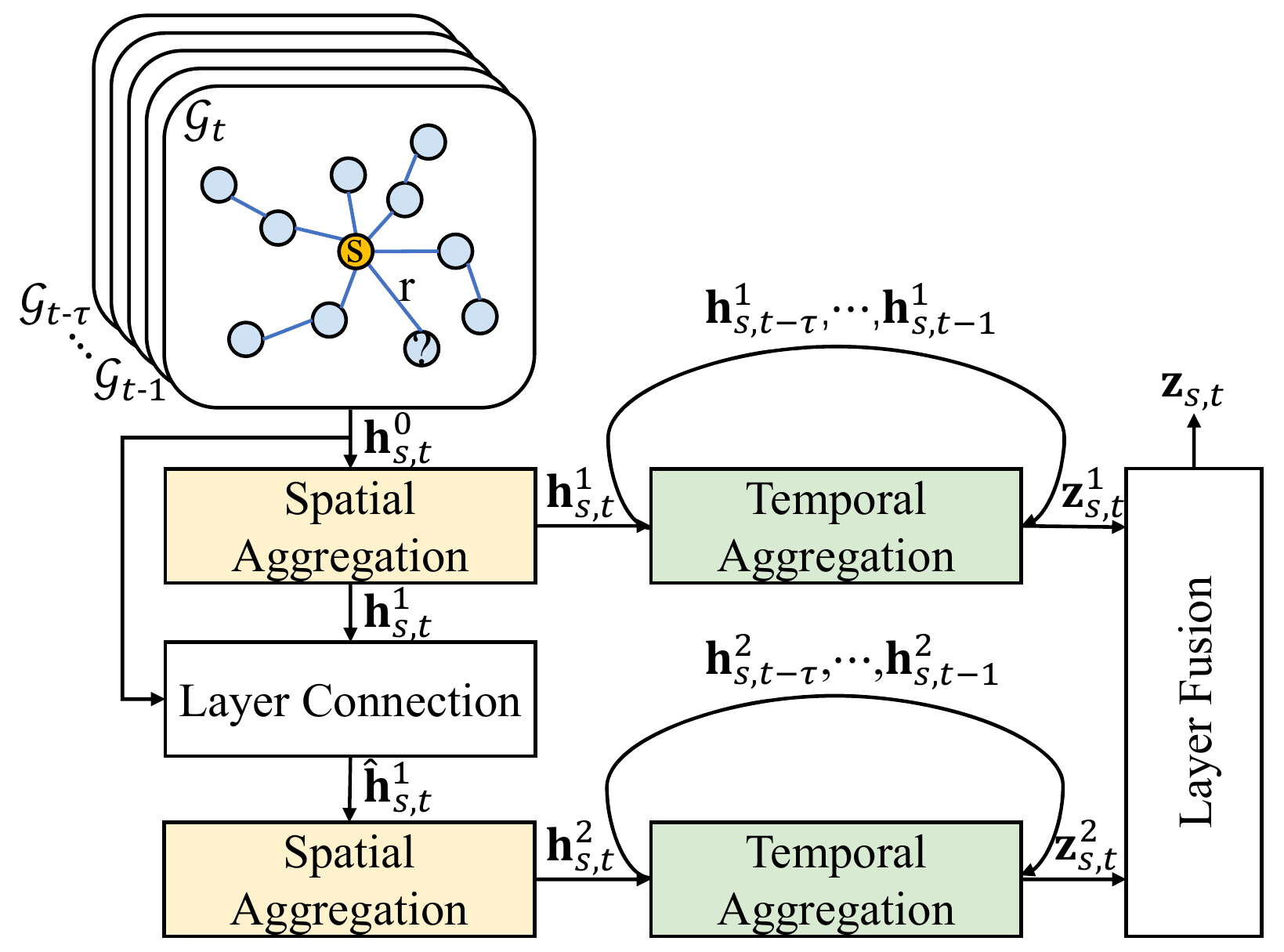}
	\caption{An illustration of the 2-layer framework. 
		The temporal aggregation module is placed after each spatial aggregation module in each layer, 
		and the layer fusion module is utilized to incorporate the intermediate feature representations produced by temporal aggregation module. 
		The layer connection module is used to help the feature reuse for each spatial aggregation.}
	\label{fig-framework1}
	\vspace{-10px}
\end{figure}

To search for data-specific and well-performing architectures based on GNN, 
we need to define a framework 
which has the ability to model topological and temporal information in TKG.
Following some existing works~\citep{taheri2019learning, sankar2020dysat, manessi2020dynamic,wu-etal-2020-temp,gao2022a,wang2022evolutionary},
we firstly discretize a TKG into multiple static KG snapshots along the time,
and utilize GNNs and sequential models to generate dynamic entity representations.
The main advantages of this approach 
include simplicity as well as 
enabling the use of a wealth of GNN and sequential model techniques.
A large number of works on temporal graphs also achieve competitive results with such this approach consisting of combinations of GNNs and recurrent architectures,
whereby the former digest graph information and the latter handle dynamism. 

Based on this motivation,
we develop a generalized framework that 
mainly consists of four key modules 
for learning expressive dynamic entity representation, 
including 
\textbf{spatial aggregation},
\textbf{temporal aggregation}, 
\textbf{layer connection}, 
and \textbf{layer fusion}. 
In Figure~\ref{fig-framework1},
we use a 2-layer architecture 
as an illustrative example of the generalized framework. 
More detailed descriptions of the four modules are as follows:

\begin{enumerate}[leftmargin=*]
	\item \textbf{Spatial Aggregation} at the $i$-th layer is conducted to 
	aggregate information from the neighbors of $s$ in static snapshot $\mathcal{G}_t$ 
	and results in the intermediate representation of entity $s$, as follows,
	\begin{equation}
		\mathbf{h}_{s,t}^1=\mathcal{O}_\text{SA}(\mathcal{G}_t,\mathbf{h}_{s}^0),
	\end{equation}
	where $\mathbf{h}_{s}^0\in\mathbf{H}$ is the initialized embedding of entity $s$, 
	$\mathbf{H}$ is the representation matrix containing embeddings of entities and relations in TKG. 

	\item \textbf{Temporal Aggregation} at the $i$-th layer generates temporal feature
	$\mathbf{z}_{s,t}^i$
	based historical feature sequences 
	$\mathbf{h}_{s,t-\tau}^i,\cdots, \mathbf{h}_{s,t-1}^i$ behind, as follows,
	\begin{equation}
		\mathbf{z}_{t}^1=\mathcal{O}_\text{TA}(\mathbf{h}_{s,t-\tau}^1,\cdots, \mathbf{h}_{s,t-1}^1, \mathbf{h}_{s,t}^1),
	\end{equation}
	where $\tau$ is a hyper-parameter, stands for the number of input KG snapshots to the model. 
	\item \textbf{Layer Connection} combines $\mathbf{h}_{s,t}^{i-1}$ with $\mathbf{h}_{s,t}^i$ to form a new representation $\hat{\mathbf{h}}_{s,t}^i$, as follows,
	\begin{equation}
		\hat{\mathbf{h}}_{s,t}^1=\mathcal{O}_\text{LC}(\mathbf{h}_{s}^0,\mathbf{h}_{s,t}^1).
	\end{equation}
	\item \textbf{Layer Fusion} generates the final representation of entity
	$\mathbf{z}_{s,t}$ 
	by fusing temporal features from temporal aggregation module in different layer, as follows,
	\begin{equation}
		\mathbf{z}_{s,t}=\mathcal{O}_\text{LF}(\mathbf{z}_{s,t}^1,\mathbf{z}_{s,t}^2).
	\end{equation}
\end{enumerate}

Based on this generalized framework, 
we can search the specific form of each operation 
to obtain data-specific architecture.
An effective search space can be naturally designed by 
including human-designed operations, 
the details of which are given in Table~\ref{tab-search-space}.

\subsection{Search Space}\label{sec-search-space}

\begin{figure}[t]
	\centering
	\includegraphics[width=0.76\linewidth]{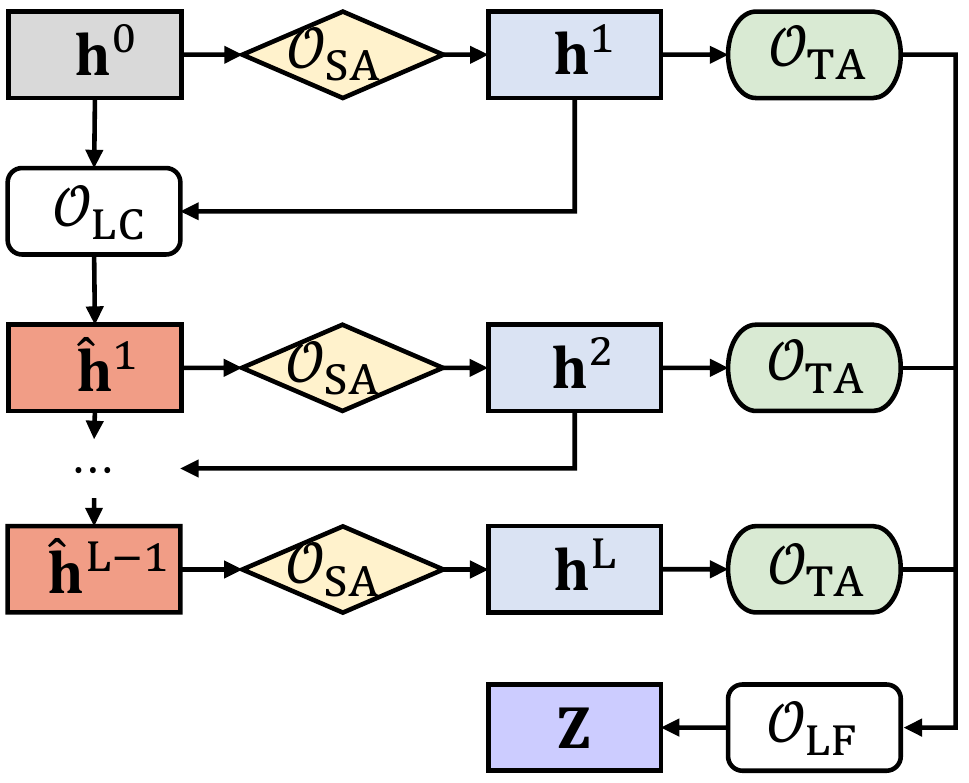}
	\caption{The illustration of search space of SPA.}
	\label{fig-search_space}
	\vspace{-8px}
\end{figure}

Based on above framework, 
we design one novel search space with a set of candidate operations 
as shown in Table~\ref{tab-search-space}. 
In the following, 
we will describe the details of these operations.

\noindent{\textbf{Spatial Aggregation.}} We choose three widely used multi-relational GNNs as alternative spatial aggregation module: RGCN \citep{schlichtkrull2018modeling}, RGAT \citep{busbridge2019relational}, CompGCN \citep{vashishth2020composition}, which denoted as \texttt{RGCN}, \texttt{RGAT}, \texttt{COMPGCN}.

\begin{table}[tb]
	\centering
	\setlength\tabcolsep{3.5pt}
	\begin{tabular}{cc}
		\toprule
		\textbf{Module name} & \textbf{Operations}               \\
		\midrule
		\makecell[c]{Spatial Aggregation  \\ $(\mathcal{O}_\text{SA})$}  & \makecell[c]{\texttt{RGCN}, \texttt{RGAT},  \\ \texttt{COMPGCN}}                              \\
		\makecell[c]{Temporal Aggregation \\$(\mathcal{O}_\text{TA}$)} & \makecell[c]{\texttt{GRU}, \texttt{SA}, \\ \texttt{IDENTITY}}                                \\
		\makecell[c]{Layer Connection \\$(\mathcal{O}_\text{LC})$}     & \makecell[c]{\texttt{LC\_SKIP}, \texttt{LC\_SUM}, \\ \texttt{LC\_CONCAT}}                    \\
		\makecell[c]{Layer Fusion \\$(\mathcal{O}_\text{LF})$}         & \makecell[c]{\texttt{LF\_MAX}, \texttt{LF\_CONCAT},\\ \texttt{LF\_SKIP}, \texttt{LF\_MEAN}} \\
		\bottomrule
	\end{tabular}
	\caption{The operations used in our search space.}
	\label{tab-search-space}
	\vspace{-8px}
\end{table}

\noindent{\textbf{Temporal Aggregation.}} For the temporal aggregation module, 
we consider two sequential models to learn temporal patterns: GRU \citep{cho-etal-2014-learning} , Self-Attention (SA) \citep{vaswani2017attention}. 
Besides, 
we incorporate the operation IDENTITY, 
which means using the results of spatial aggregation directly, 
i.e., $\mathbf{z}_{s,t}^i=\mathbf{h}_{s,t}^i$, 
rather than learning dynamic feature between snapshots. 

\noindent{\textbf{Layer Connection.}} It has been well proven in many literatures \citep{li2021deepgcns} that the use of skip connections between spatial aggregation modules can help alleviate over-smoothing and the vanishing gradient issue, 
and improve the performance of the model. 
In our search space, 
we add three different skip connection operations to encourage various feature reuse, 
i.e., \texttt{LC\_SKIP}, \texttt{LC\_SUM}, \texttt{LC\_CONCAT}. 

\noindent{\textbf{Layer Fusion.}} In static graph learning, 
some studies~\citep{xu2018representation} 
focus on obtaining more expressive structure-aware representation 
by selectively fusing the intermediate representation of spatial aggregation. 
We borrow this idea to temporal graph learning and 
provide four fusion operations 
to integrate the representations of the intermediate temporal aggregation layers with 
the average, 
maximum, 
concatenation 
and skip, 
denoted as \texttt{LF\_MEAN}, \texttt{LF\_MAX}, \texttt{LF\_CONCAT} and \texttt{LF\_SKIP}, 
respectively. 
The search for various fusion operations 
allows the model to learn to adapt to different dynamic subgraph structures. 
 
An example of the L-layer search space is shown in Figure~\ref{fig-search_space}.
With so many candidate architectures in the search space, 
SPA can use efficient search algorithm 
to obtain data-specific architectures 
beyond existing human-designed ones.

\subsection{Search Objective}

Let the training and validation set be 
$\mathcal{D}_\text{tra}$ and $\mathcal{D}_\text{val}$,
$\mathcal{N}(\bW_{\Theta,\mathbf{H}};\bal)$ be a TKGC model 
(where $\bW_{\Theta,\mathbf{H}}$ represents model parameters containing model weights $\Theta$ and TKG embedding $\mathbf{H}$,
and $\bal$ is the model architecture),
$\mathcal{M}$ be the measurement on $\mathcal{D}_\text{val}$
and $\mathcal{L}$ be the loss on $\mathcal{D}_\text{tra}$.
The problem is defined to find an architecture $\bal$ 
such that validation performance is maximized, i.e.,
\begin{align}
	\bal^* &= \argmax\nolimits_{\bal \in \mathcal{A}}\;
	\mathcal{M} (\mathcal{N}(\bW_{\Theta,\mathbf{H}}^*;\bal), \mathcal{D}_\text{val}),
	\label{eq-nested-nas-opt}
	\\\
	\text{\;s.t.\;}&\bW_{\Theta,\mathbf{H}}^*
	= \argmin\nolimits_\bW \mathcal{L}(\mathcal{N}(\bW_{\Theta,\mathbf{H}};\bal), \mathcal{D}_\text{tra}),
	\label{eq-nested-nas-opt:2}
\end{align}
which is a bi-level optimization problem and is non-trivial to solve. 
Because the computation cost to get the optimal parameters $\bW_{\Theta,\mathbf{H}}^*$ is generally high.
And the search space is large.
Thus, 
how to efficiently search the architectures is a big challenge.

To perform TKGC task, 
we use score function to measure the plausibility of each candidate quadruple $(s, r, o, t)$.
Since our proposed framework can generate time-aware entity embeddings,
we only need static score function.

Specifically, 
the score function for quadruple is defined as follows in SPA:
\begin{equation}\label{equ-decoder}
	\phi(s, r, o, t)= f(\bm{z}_{s, t}, \bm{h}_r, \bm{z}_{o, t}),
\end{equation}
where $\bm{z}_{s, t}$ and $\bm{z}_{o, t}$ are time-aware representaions for subject and object entities, while $\bm{z}_r$ is a learned embedding of the relation $r$. 
In this work, 
we use ComplEx \citep{trouillon2016complex} as the score function, 
which is known to perform well on static KGC benchmarks.

For the loss function,
following the setting of TeMP,
we employ the cross-entropy loss for parameter learning.
More details about loss function is in the Appendix~\ref{sec:appendix-loss}.

\subsection{Search Algorithm}\label{sec-search-algorithm}
 
\begin{algorithm}[tb]
	\caption{SPA - Search to PAss messages}
	\begin{algorithmic}[1]
		\Require{Training dataset $\mathcal{D}_{\text{tra}}$, 
			validation dataset $\mathcal{D}_{\text{val}}$, 
			the epoch $T_1$ for train supernet, the epoch $T_2$ for search architecture, the search space $\mathcal{A}$.}
		\Ensure{The searched architecture.}
		\State Random initialize the parameter of supernet $\mathbf{W}$.
		\While {$t<T_1$}
		\For {each minibatch $\mathcal{B} \in \mathcal{D}_{\text{train}}$}
		\State Random sample $\bal$ from $\mathcal{A}$.
		\State Calculate the training loss $\mathcal{L}_{\text{tra}}$ for $\bal$.
		\State Update weight subset $\mathbf{W}_{\Theta,\mathbf{H}}(\bal)$ with $\mathcal{L}_{\text{tra}}$.
		\EndFor
		\EndWhile
		\While {$t<T_2$}
		\State Random sample $\bal$ from $\mathcal{A}$.
		\State Inherit weight subset $\mathbf{W}_{\Theta,\mathbf{H}}^*(\bal)$ from $\mathbf{W}_{\Theta,\mathbf{H}}^*$.
		\State Calculate the validation performance for $\bal$ in $\mathcal{D}_{\text{val}}$.
		\EndWhile \\
		\Return The searched architecture with the highest validation performance.
	\end{algorithmic}
	\label{alg-spa}
	\vspace{-2px}
\end{algorithm}

Based on the proposed framework and the search space, 
the search algorithm is used to search operations from the corresponding operation set.

Inspired by recent advances in NAS, 
we propose to solve Equation~\eqref{eq-nested-nas-opt}, \eqref{eq-nested-nas-opt:2} 
using one-shot NAS paradigm, 
which greatly improves the efficiency of performance estimation by training only one supernet.

There are two types of methods in one-shot NAS: 
the single-stage method and the two-stage method.
The first one combines supernet training and search in a \emph{single stage}.
Representative methods include DARTS~\citep{liu2019darts}, SNAS~\citep{xie2019snas}, etc.
The single-stage approach requires that the validation metrics be differentiable 
to allow supernet training and architecture search to be jointly optimized by gradient-based methods,
which is inappropriate for our task as its metric(i.e. MRR) is non-differentiable. 
And the correlation between the validation loss and the validation metric is unclear.
Using the validation loss to update the architecture parameters may mislead the search algorithm to find a sub-optimal architecture.

Moreover, 
the single-stage approach requires training the whole supernet, 
which demands tremendous GPU memory as the proposed search space contains spatial encoder and temporal encoder. 
Hence,
we adopt the two-stage approach,
which decouples supernet training and architecture search.
 
In this paper, 
we adopt SPOS~\citep{guo2020single},
a typical two-stage method,
as it can consume the GPU memory less and fully train each candidate operation.
Algorithm \ref{alg-spa} delineates the full procedure. 

\begin{figure}[t]
	\centering
	\includegraphics[width=1\linewidth]{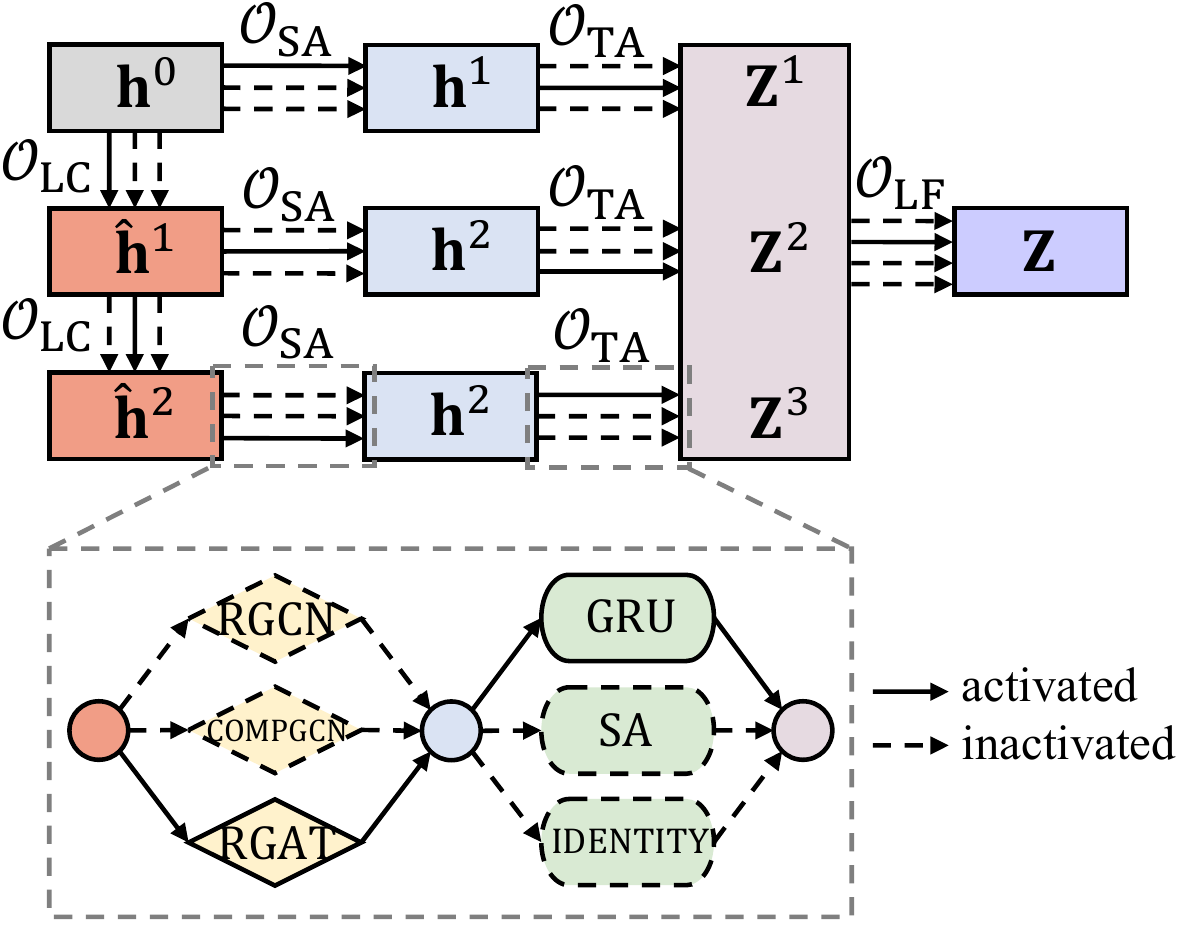}
	\caption{The illustration of the single path supernet.
		In the training stage, 
		the weights of the solid line part (\texttt{RGAT}, \texttt{GRU}) are activated and updated, the dotted portions are masked and inactivated.}
	\label{fig-search_algorithm}
	\vspace{-10px}
\end{figure}

\noindent{\textbf{Supernet Training.}} For the solution of Equation~\eqref{eq-nested-nas-opt:2}, 
following SPOS,
we construct a supernet structure 
that each candidate architecture is a single path.
In each step of optimization, 
as shown in Figure~\ref{fig-search_algorithm}.
an architecture $\bal$ (one path, i.e., the solid part of the figure) is sampled from the search space 
in a uniformly distributed manner.
It guarantees equal expectations of the number of times each architecture is sampled, 
thus all architectures (and their weights) are trained fully and equally.
And then, 
only the weights corresponding $\bal$ are activated and updated.
So the GPU memory usage is efficient. 

\noindent{\textbf{Architecture Search.}} After getting the trained optimal weights of supernet $\bW_{\Theta,\mathbf{H}}^*$, 
to solve the problem in Equation~\eqref{eq-nested-nas-opt}, 
we leverage random search to find well-performing architecture $\bal$. 
This is simple but effective for our search space. 

Finally, 
architecture with the highest validation performance(i.e. validation MRR) in all iterations will be returned. 

\section{Experiments}

\subsection{Experimental Settings}

\noindent\textbf{Datasets.} We perform evaluation on three widely used TKG completion datasets, including ICEWS14~\citep{garcia-duran-etal-2018-learning}, ICEWS05-15~\citep{garcia-duran-etal-2018-learning} and GDELT~\citep{leetaru2013gdelt}.
ICEWS14 and ICEWS05-15 are two subsets of \textit{Integrated Crisis Early Warning System} (ICEWS) database with different time spans.
GDELT is a subset of \textit{Global Database of Events, Language, and Tone} (GDELT), 
which contains facts facts from April 1, 2015 to March 31, 2016.
The detailed dataset statistics is presented in the Appendix~\ref{sec:appendix-dataset}.

\noindent\textbf{Evaluation Metrics.} We follow~\citep{bordes2013translating} to use the filtered ranking-based metrics, i.e., mean reciprocal ranking (MRR) and Hit@1/3/10 for evaluation.
For both metrics, 
the larger value indicates the better performance.

\noindent\textbf{Baseline Methods.} We compare SPA with two types of baselines: human-designed methods and NAS methods. 

For human-designed methods, 
we take TransE~\citep{bordes2013translating}, 
Distmult~\citep{yang2015embedding}, 
ComplEx~\citep{trouillon2016complex} and SimplE~\citep{kazemi2018simple} to represent static KG completion methods, and TTransE~\citep{leblay2018deriving}, 
TA-Distmult~\citep{garcia-duran-etal-2018-learning}, 
HyTE~\citep{dasgupta-etal-2018-hyte}, 
DE-SimplE~\citep{goel2020diachronic}, 
TNTComplEx~\citep{lacroix2020tensor}, 
ChronoR~\cite{sadeghian2021chronor}, 
TeLM~\citep{xu2021temporal} and
BoxTE~\citep{messner2022temporal} to represent state-of-the-art embedding-based methods designed for TKGC.
For the GNN-based methods, 
we compare with both TeMP~\citep{wu-etal-2020-temp} and T-GAP~\citep{jung2021learning} here.

For NAS methods, 
since existing methods cannot learn the data-specific architecture for temporal graph, 
we further provide Random search as the baseline for comparisons based on the proposed search space in Section~\ref{sec-search-space}.

\noindent\textbf{Implementation and Hyperparameters.} For all NAS methods (Random baseline and SPA), 
we derived the candidate GNNs from the search space in the search process. 
All the searched candidates are tuned individually with hyperparameters like learning rate, weight decay, etc.
In this paper, 
the 3-layer framework is empirically chosen for all NAS methods on all datasets.
We set the negative sampling ratio to 500, i.e. 500 negative samples per positive triple.
More details about the implementation and hyperparamters are given in Appendix~\ref{sec:appendix-impl}.

\subsection{Performance Comparison}

\begin{table*}[ht]
	\centering 
	\setlength\tabcolsep{2.5pt}
	\scalebox{0.93}{
	\begin{tabular}{cccccccccccccc}
		\toprule
		\multirow{2}*{\textbf{Type}} &\multirow{2}*{\textbf{Model}}              &\multicolumn{4}{c}{\textbf{ICEWS14}}   & \multicolumn{4}{c}{\textbf{ICEWS05-15}}      & \multicolumn{4}{c}{\textbf{GDELT}}\\
		\cmidrule(lr){3-6} \cmidrule(lr){7-10} \cmidrule(lr){11-14}
		&&\textbf{MRR} & \textbf{H@1} &\textbf{H@3} &\textbf{H@10}
		&\textbf{MRR} & \textbf{H@1} &\textbf{H@3} &\textbf{H@10}
		&\textbf{MRR} & \textbf{H@1} &\textbf{H@3} &\textbf{H@10}\\
		\midrule
		\multirow{16}{*}{\shortstack{Human-\\designed}}&TransE  &0.326 &15.4 &43.0 &64.4 &0.330 &15.2 &44.0 &66.0 &0.155 &6.0 &17.8 &33.5  \\
		&DistMult    &0.441 &32.5 &49.8 &66.8 &0.457 &33.8 &51.5 &69.1 &0.210 &13.3 &22.4 & 36.5 \\
		&ComplEx     &0.442 &40.0 &43.0 &66.4 &0.464 &34.7 &52.4 &69.6 &0.213 &13.3 &22.5 &36.6  \\
		&SimplE      &0.458 &34.1 &51.6 &68.7 &0.478 &35.9 &53.9 &70.8 &0.206 &12.4 &22.0 &36.6 \\
		\cline{2-14}
		&TTransE     &0.255 &7.4  &-    &60.1 &0.271 &8.4  &-    &61.6 &0.115 &0.0  &16.0 &31.8 \\
		&HyTE        &0.297 &10.8 &41.6 &65.5 &0.316 &11.6 &44.5 &68.1 &0.118 &0.0  &16.5 &32.6 \\
		&TA-DistMult &0.477 &36.3 &-    &68.6 &0.474 &34.6 &-    &72.8 &0.206 &12.4 &21.9 &36.5 \\
		&DE-SimplE   &0.526  &41.8   &59.2   &72.5   &0.513  &39.2   &57.8   &74.8  &0.230 &14.1 &24.8 &40.3  \\
		&TNTComplEx  &0.620 &52.0   &66.0   &76.0   &0.670  &59.0   &71.0   &81.0  &-     &-    &-    &-    \\
		&TIMEPLEX 	 &0.604	&51.5	&-	    &77.1	&0.640	&54.5   &-		&81.8  &-     &-    &-    &-    \\	
		& ChronoR	 &0.625	&\textbf{54.7}	&66.9	&77.3	&0.675	&\underline{59.6}	&72.3	&82.0  &-     &-    &-    &-    \\				
		&TeLM	     &0.625	&\underline{54.5}	&67.3	&77.4	&0.678	&\textbf{59.9}	&72.8	&82.3  &-     &-    &-    &-    \\				
		&BoxTE	     &0.613	&52.8	&66.4	&76.3	&0.667	&58.2	&71.9	&82.0  &0.352 &26.9 &37.7 &\textbf{51.1} \\ \cline{2-14}
		&TeMP-GRU    &0.601 &47.8   &68.1   &82.8   &0.691  &56.6   &78.2   &\underline{91.7}  &0.275 &19.1 &29.7 &43.7 \\
		&TeMP-SA     &0.607 &48.4   &68.4   &84.0   &0.680  &55.3   &76.9   &91.3  &0.232 &15.2 &24.5 &37.7 \\
		&T-GAP	     &0.610	&50.9	&67.7	&79.0	&0.670	&56.8	&74.3	&84.5  &-     &-    &-    &-    \\
		
		\hline
		\multirow{2}{*}{NAS} & Random & \underline{0.642} &52.8    &\underline{72.2} &\underline{84.3}          &    \underline{0.701}          &58.0 & \underline{78.8}& \underline{91.7} & \underline{0.353} &\underline{27.1} & \underline{37.9} & \textbf{51.1}   \\
		& \textbf{SPA}   & \textbf{0.658} &54.4 &\textbf{73.7} &\textbf{85.7} & \textbf{0.713} & 58.0 & \textbf{82.0} & \textbf{93.3} &\textbf{0.360}	&\textbf{28.2} &\textbf{38.4}	&\underline{51.0}      \\
		\bottomrule
		
	\end{tabular}
	}
	\caption{Temporal KG completion evaluation results on ICEWS14, ICEWS05-15 and GDELT. 
	The H@1, H@3, and H@10 metrics are multiplied by 100. 
	Best results are in bold and the second best is underlined. 
	"-" means that results are not reported in those papers or their code on that data/metric is not available.} 
	\label{tab-result}
	\vspace{-10px}
\end{table*}

Table~\ref{tab-result} shows the overall result on three benchmarks.
As can be seen, there is no clear winner among the human-designed baselines on all datasets.
Besides, 
we can see that SPA consistently outperforms all baselines on all datasets, 
which demonstrates the effectiveness of SPA on searching for data-specific architectures for TKGC.

When it comes to NAS baselines, 
the performance gains of SPA are also significant. 
On one hand, 
the Random baselines achieve considerable performance gains on all these datasets, 
which demonstrates the effectiveness of the search space.
On the other hand, 
compared with Random, 
which use the designed search space of SPA, 
the performance gains are from the single path one-shot search algorithm on obtaining better architectures.

\begin{figure}[!ht]
	\vspace{-10px}
	\subfloat{
		\begin{minipage}[t]{0.5\linewidth}
			\centering
			\includegraphics[width=0.95\linewidth]{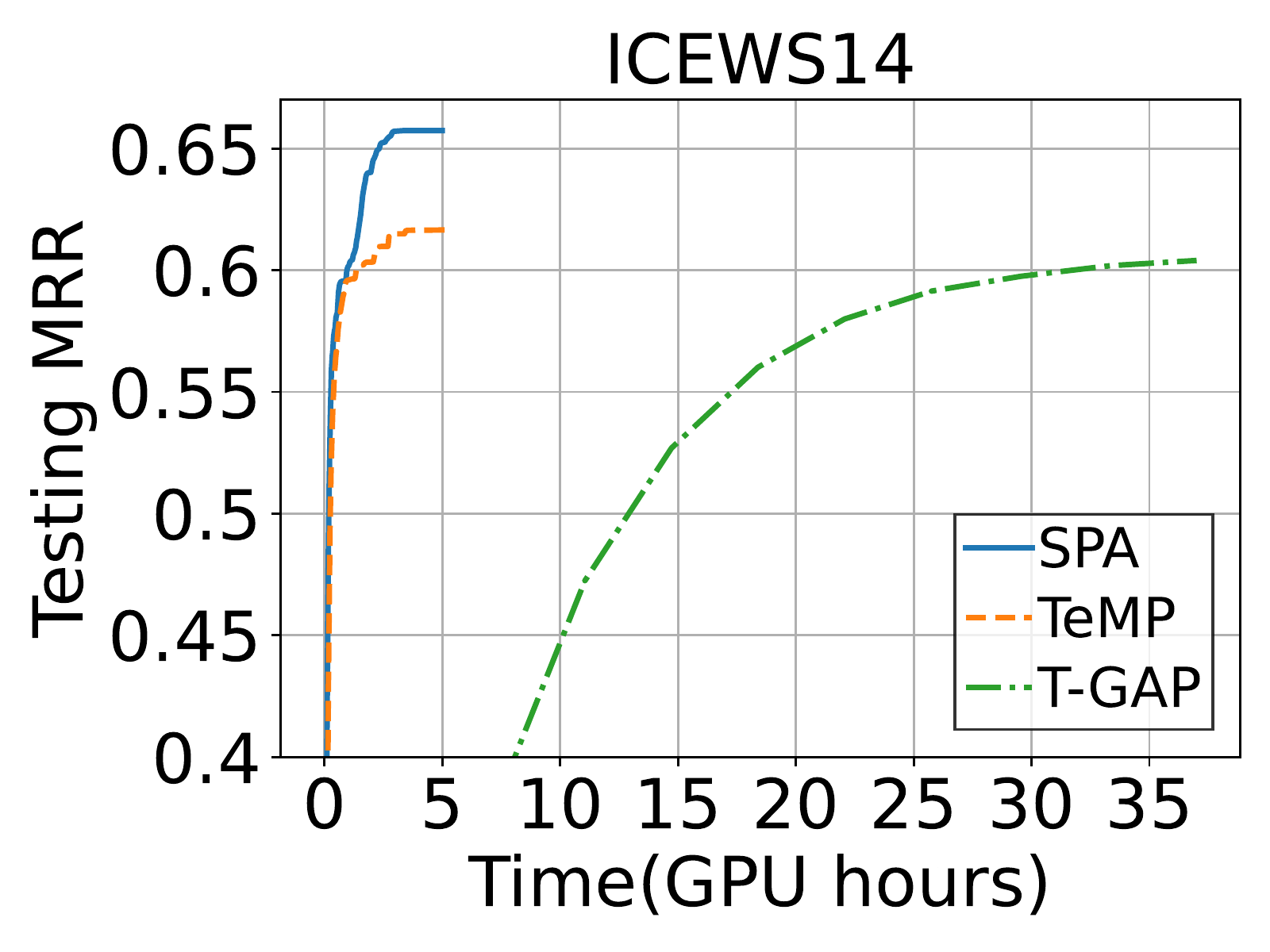}
		\end{minipage}%
	}%
	\subfloat{
		\begin{minipage}[t]{0.5\linewidth}
			\centering
			\includegraphics[width=0.95\linewidth]{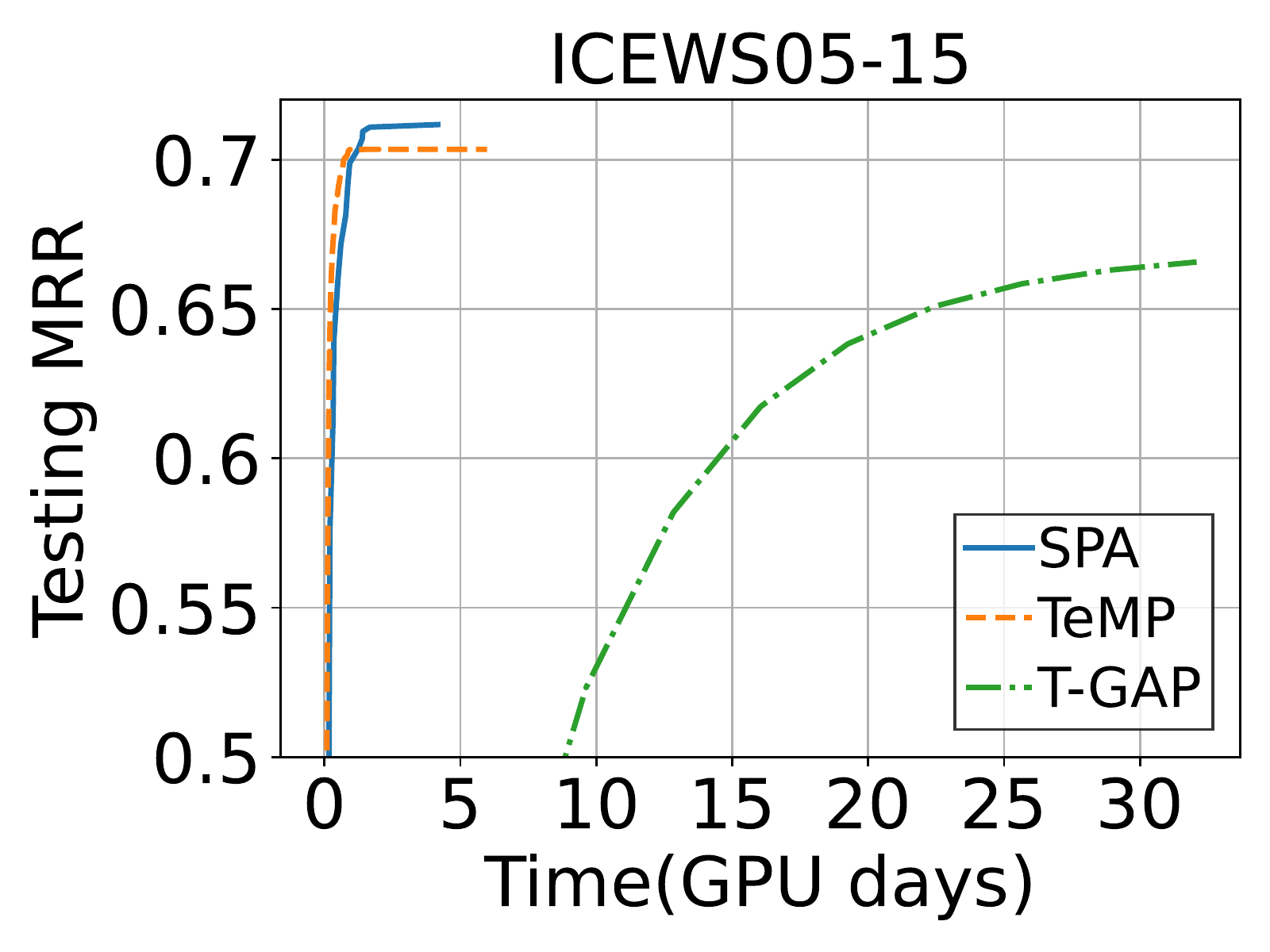}
		\end{minipage}%
	}
	\caption{Comparison on convergence between the searched architectures (by SPA) and human-designed GNN-based methods.} 
	\label{fig-learning_curve}
	\vspace{-10px}
\end{figure}

Figure~\ref{fig-learning_curve} shows the learning curves of GNN-based methods on ICEWS14 and ICEWS05-15, 
including TeMP, 
T-GAP and the proposed SPA.
As can be seen,
the searched architecture not only outperform baselines, 
but also have comparable time as the other GNN-based methods,
which demonstrates the searched architecture can better capture diverse topological and temporal properties of different TKGs.

Further, 
we visualize the searched architectures on three benchmark datasets in Figure~\ref{fig-searched-archs}, 
from which it is clear that different operation combinations of these four modules 
are obtained, 
i.e., data-specific architectures.
We will discuss the details about the searched architectures in Section~\ref{sec:case-study}.

Therefore, 
these results demonstrate the need for data-specific methods for TKGC, 
and at the same time, 
the effectiveness of SPA on designing adaptive architectures.

\subsection{Understanding the Search Algorithm}

In this part,
we evaluate the search algorithm from the perspectives of the efficiency of search algorithm, 
the effectiveness of weight sharing, 
and the choice of validation metric. 

\subsubsection{Efficiency of Search Algorithm}

\begin{figure}[h]
	\vspace{-10px}
	\subfloat{
		\begin{minipage}[t]{0.5\linewidth}
			\centering
			\includegraphics[width=0.95\linewidth]{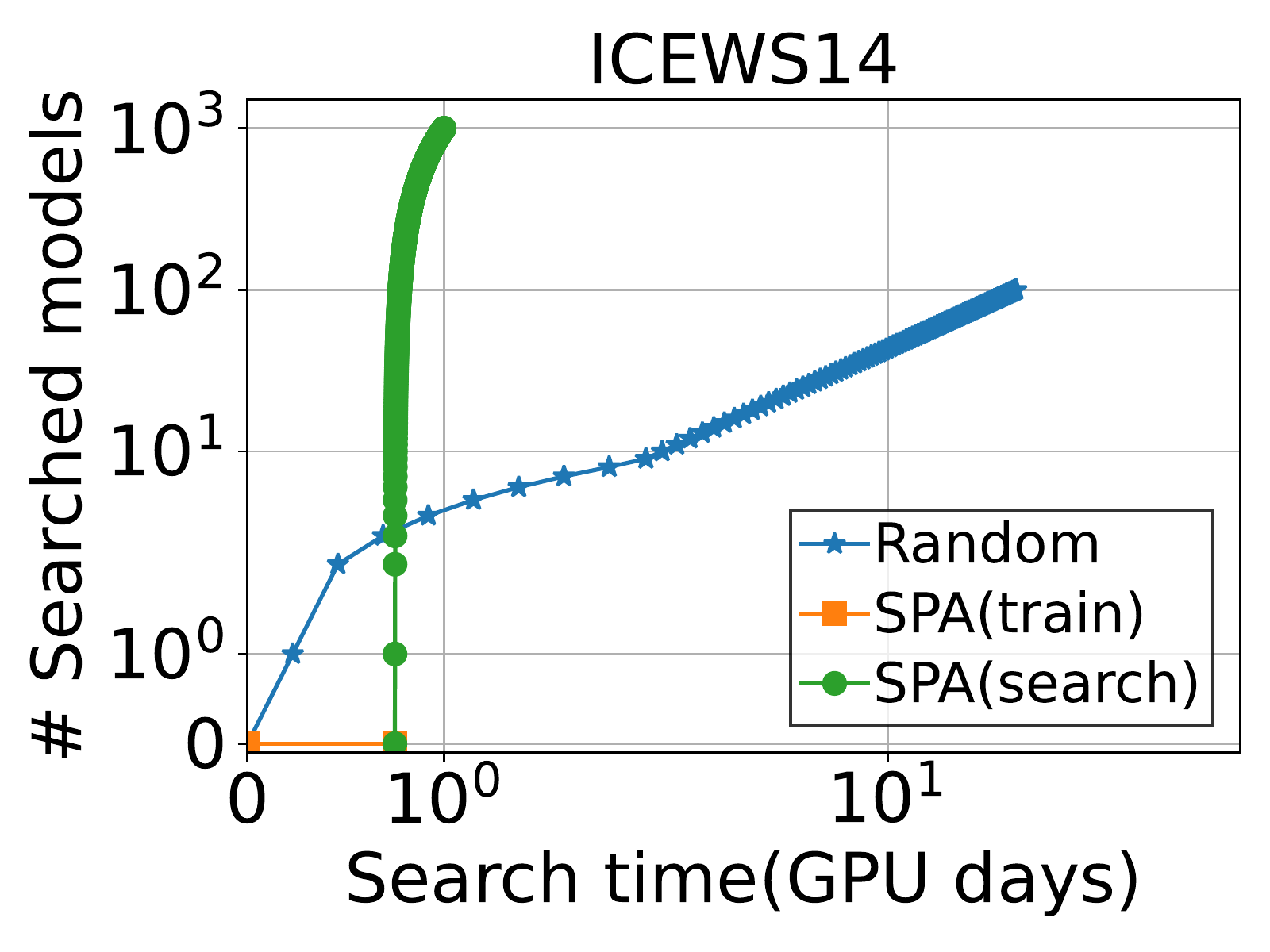}
		\end{minipage}%
	}%
	\subfloat{
		\begin{minipage}[t]{0.5\linewidth}
			\centering
			\includegraphics[width=0.95\linewidth]{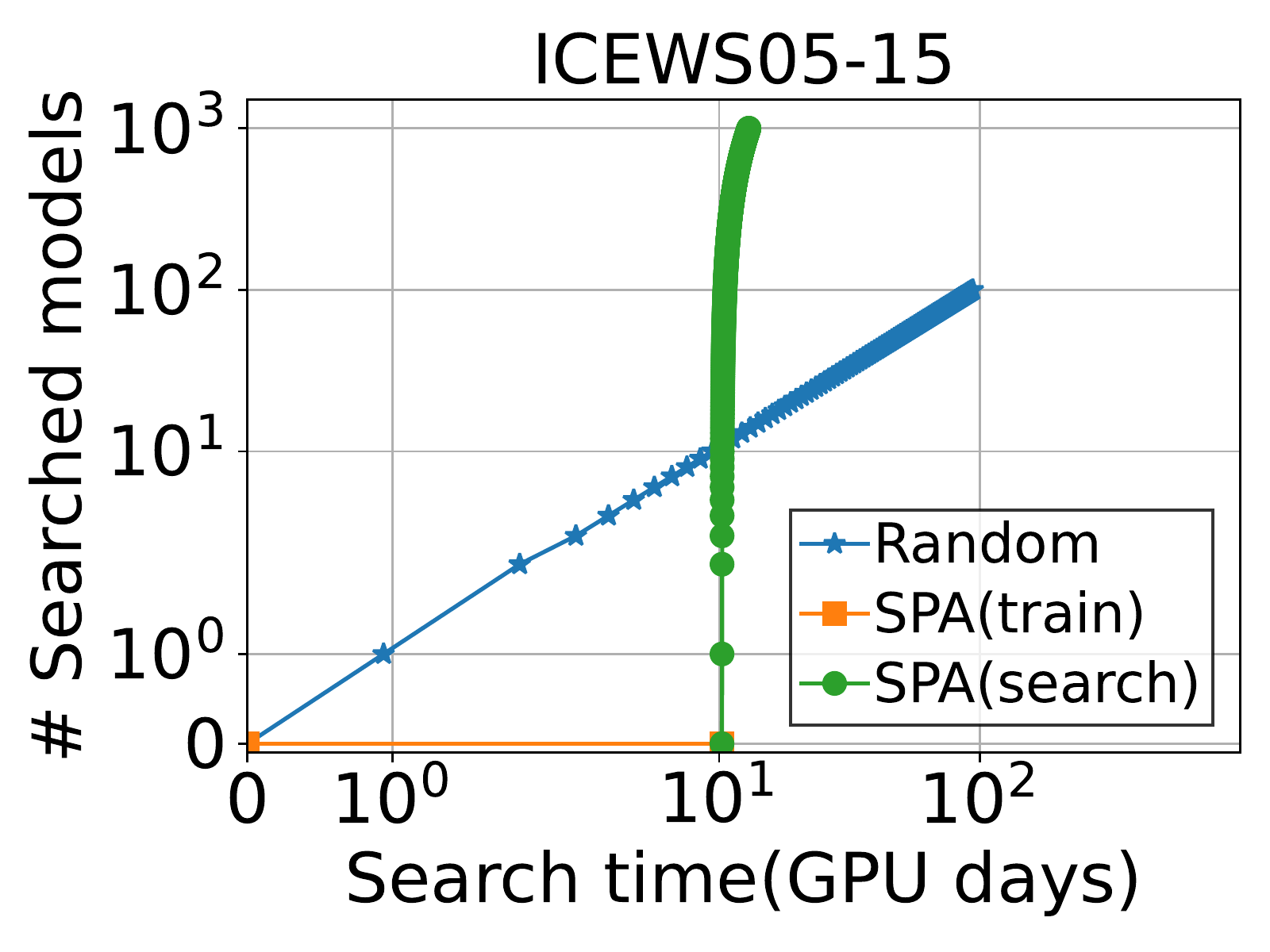}
		\end{minipage}%
	}
	\caption{Comparison of SPA with Random search during the search process.} 
	\label{fig-search_time}
	\vspace{-10px}
\end{figure}

To show the efficiency of the search algorithm,
we compare SPA with Random search baseline.
Figure~\ref{fig-search_time} shows the variation in the number of searched models during the search process.

As can be seen,
random search have to take a long time 
to train each candidate architecture from scratch,
while SPA spend most of the time on training supernet.
In the stage of architecture search,
SPA directly picks the corresponding weights 
from the trained supernet 
for the specific architecture evaluation,
which significantly improves the efficiency compared to random search.
This is mainly attributed to the weight-sharing strategy.
\subsubsection{Effectiveness of Weight Sharing}

\begin{figure}[ht]
	\centering
	\includegraphics[width=0.9\linewidth]{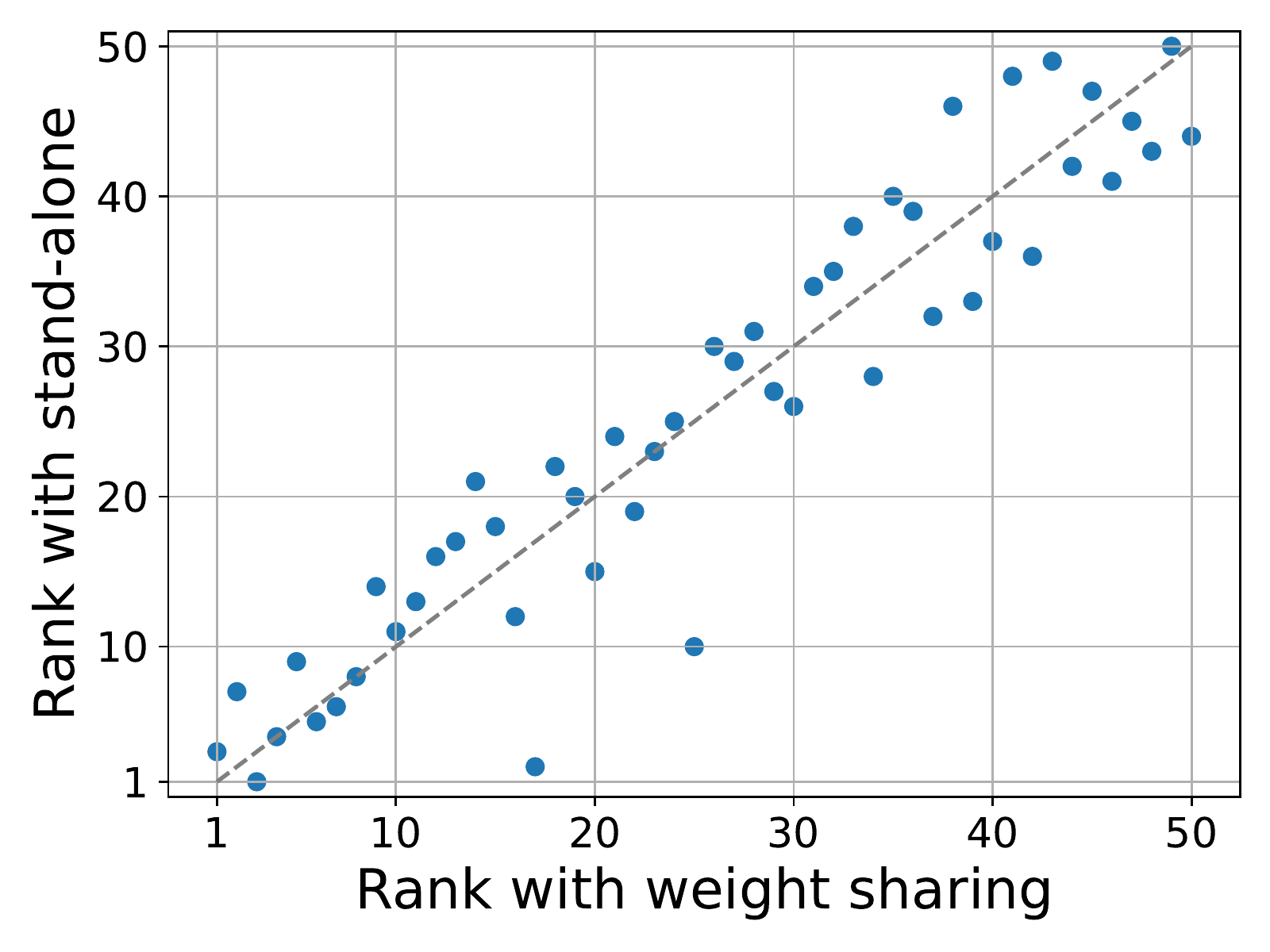}
	\caption{Rank correlation between stand-alone and weight sharing approach.}
	\label{fig-corr}
	\vspace{-10px}
\end{figure}

To demonstrate the effectiveness of weight sharing, 
we empirically visualize the rank correlation of the validation performance
between the weight sharing strategy
and the stand-alone approach,
as shown in Figure~\ref{fig-corr}. 
For the stand-alone approach,
we randomly sample 50 architectures $\mathcal{C}$, 
train and evaluate them from scratch.
About weight sharing,
we inherit the corresponding subweights of the trained supernet for each structure in $\mathcal{C}$ and evaluate it.

As can be seen, 
it is obvious that 
the rank of weight sharing validation MRR 
has near positive correlation 
with the rank of stand-alone validation MRR.
And then, 
most structures that have high estimated ranks by weight sharing 
truly have high ranks 
using the setting of stand-alone.
This demonstrates that the weight sharing strategy can search for good structures.

\subsubsection{Choice of Validation Metric}

In Section~\ref{sec-search-algorithm}, 
we discuss the rationality of adopting the SPOS~\citep{guo2020single} method for search algorithm.
Here,
to show the impact of validation metric for SPOS, 
we compare the following SPA variants: 
(i) SPA(train loss), 
which uses training loss rather than valid MRR for evaluating candidate architecture in the stage of architecture search; 
(ii) SPA(valid loss), 
which uses validation loss for evaluating candidate architecture. 
Moreover, 
we adopt two variants of DARTS~\citep{liu2019darts} as search algorithms, 
including SPA-D(train loss) and SPA-D(valid loss), 
which use gradient-based optimization to update architecture parameters by minimizing training loss and validation loss, 
respectively.

Table~\ref{tb-metric-evaluation} shows the testing MRRs of different variants on ICEWS14 and GDELT.
As can be seen,
the use of validation MRR can help to select the better sub-network.
The variants associated with DARTS run out of memory on GDELT with 3 million facts due to the demand for tremendous GPU memory.
Besides, 
when using the same validation metric, 
the performances of architecture searched by SPOS consistently outperform that of DARTS, 
which may be due to the coupling of supernet weights and architecture parameters leading to the selection of inferior architectures.

\begin{table}[!h]
	\centering
	\setlength\tabcolsep{3pt}
	\scalebox{0.93}{
		\begin{tabular}{cccc}
			\toprule
			\makecell[c]{\textbf{Search} \\ \textbf{algorithm}} & \textbf{Variant} & \textbf{ICEWS14} & \textbf{GDELT}          \\ 
			\midrule
			\multirow{2}{*}[-1ex]{DARTS}  & SPA-D(train loss) &  0.547 &  OOM   \\
			\cmidrule{2-4}
			&           SPA-D(valid loss)            &         0.615         &          OOM         \\
			\midrule
			\multirow{3}{*}[-1ex]{SPOS} &         SPA(train loss)          &         0.587       &        0.324           \\
			\cmidrule{2-4}
			&            SPA(valid loss)            &         0.623          &             0.341     \\
			\cmidrule{2-4}
			&              \textbf{SPA(valid MRR)}              &     \textbf{0.658}     &     \textbf{0.360}     \\ 
			\bottomrule
		\end{tabular}
	}
	\caption{Performance of SPA using different variants of search algorithm. 
	"OOM" means out of memory.}
	\label{tb-metric-evaluation}
	\vspace{-10pt}
\end{table}
\begin{figure*}[!t]
	\centering
	\subfloat[ICEWS14]{
		\includegraphics[width=0.3\linewidth]{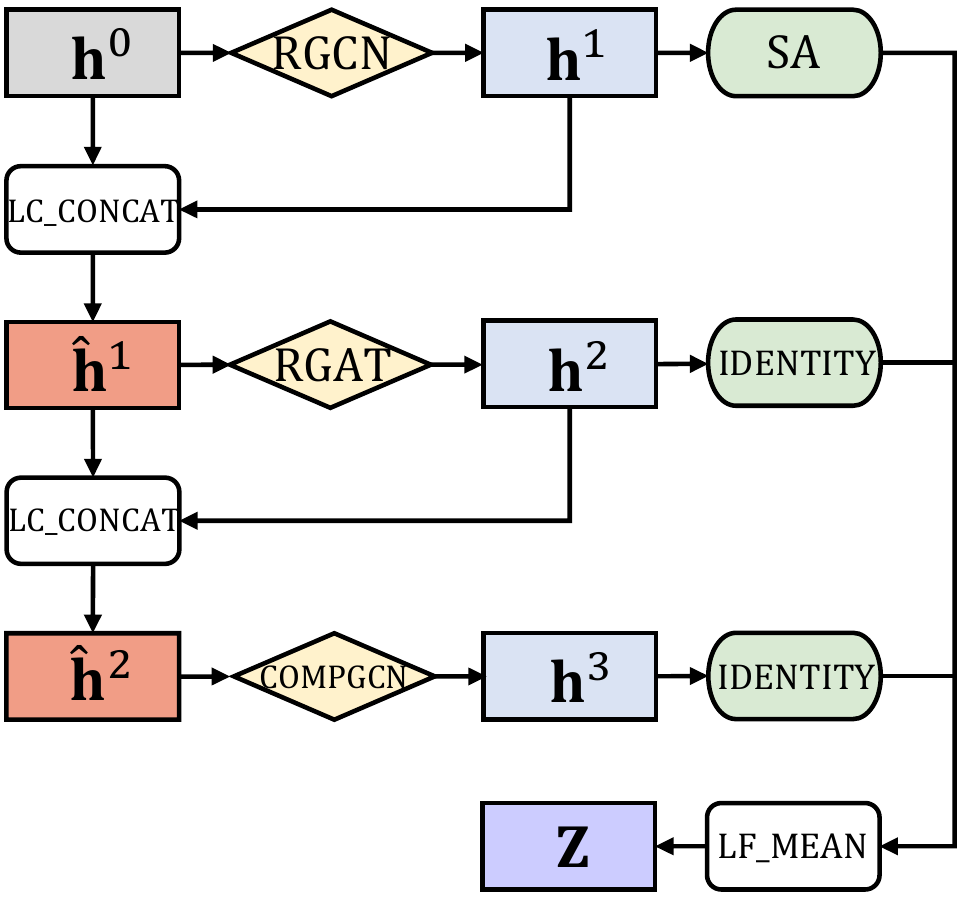}
	}%
	\subfloat[ICEWS05-15]{
		\includegraphics[width=0.3\linewidth]{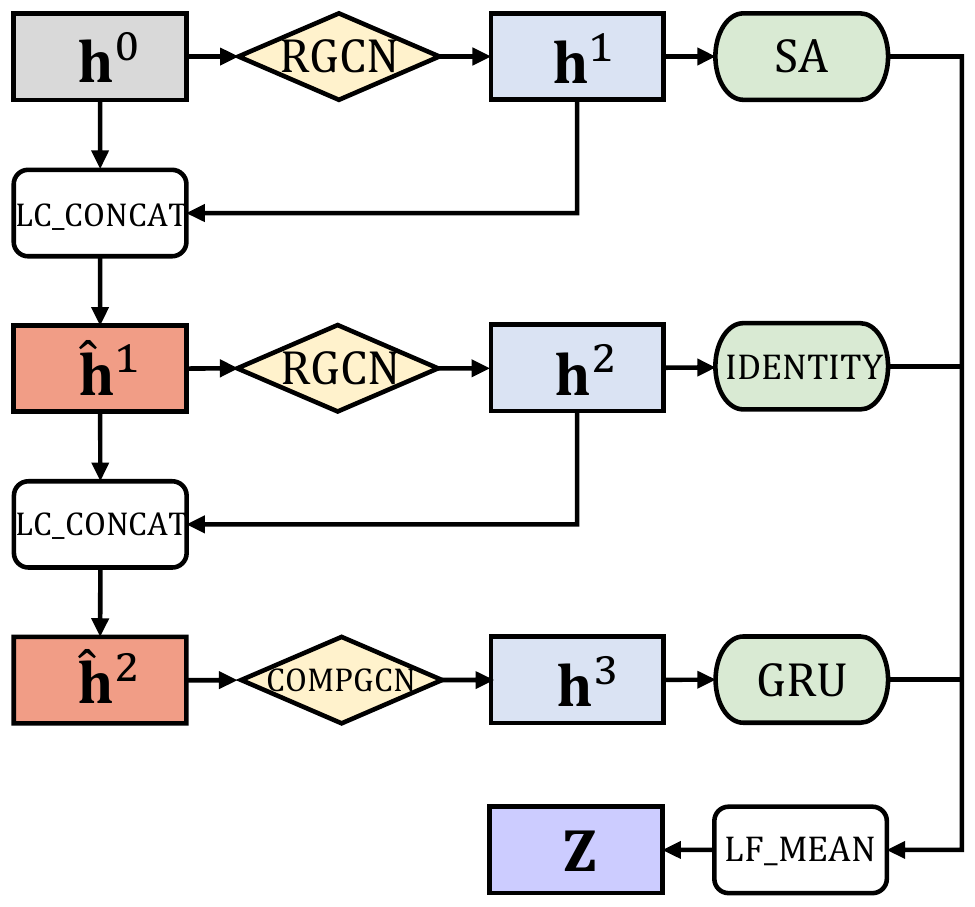}
	}
	\subfloat[GDELT]{
		\includegraphics[width=0.3\linewidth]{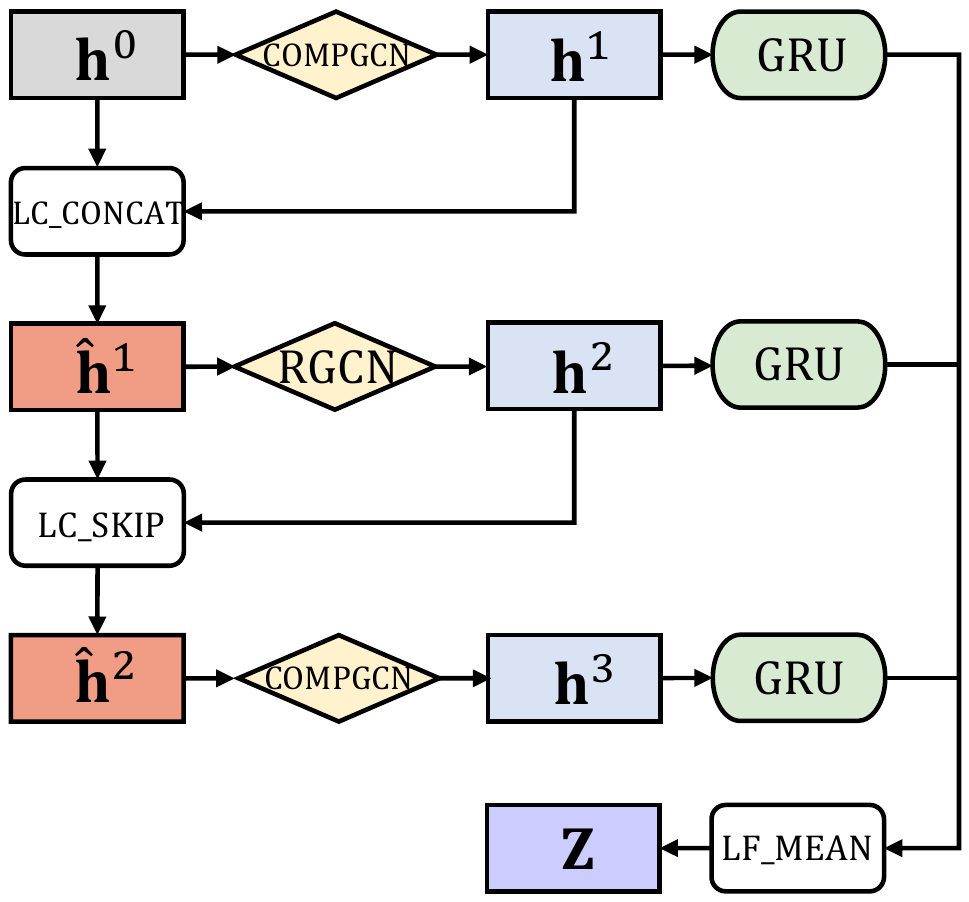}
	}
	\caption{The searched architectures on three benchmark datasets.} 
	\label{fig-searched-archs}
	\vspace{-10px}
\end{figure*}

\subsection{Ablation Studies on the Search Space}

We conduct ablation studies to show the influences of the four modules in the search space. 
For simplicity, we use two datasets: ICEWS14 and GDELT, 
and run SPA over different variants of search space, 
for which the results are shown in Table~\ref{tb-space-evaluation}.

\begin{table}[!ht]
	\centering
	\setlength\tabcolsep{1pt}
	\scalebox{0.95}{
	\begin{tabular}{cccc}
		\toprule
		\textbf{Fixed} & \textbf{Variant} & \textbf{ICEWS14} & \textbf{GDELT}          \\ 
		\midrule
		\multirow{2}{*}{\shortstack{Spatial \\Aggregation}\vspace{-2mm}}  & SPA-RGCN &  0.648 &  0.347   \\
		\cmidrule{2-4}
		                                                     &           SPA-RGAT            &         0.653         &          0.357         \\
		\midrule
		\multirow{2}{*}{\shortstack{Temporal \\ Aggregation}\vspace{-2mm}} &         SPA-IDENTITY          &         0.654       &        0.342           \\
		\cmidrule{2-4}
		                                                      &            SPA-GRU            &         0.585          &             0.358     \\
		\midrule
		\multirow{2}{*}{\shortstack{Layer \\Connection}}    & \multirow{2}{*}{SPA-LC\_SKIP} & \multirow{2}{*}{0.655} & \multirow{2}{*}{0.356} \\
		                      &     &                           &  \\
		\midrule
		                    Layer Fusion                      &         SPA-LF\_SKIP          &         0.623          &         0.349          \\ 
		\midrule
		                                                      &              \textbf{SPA}              &     \textbf{0.658}     &     \textbf{0.360}     \\ 
		\bottomrule
	\end{tabular}
	}
	\caption{Performance of SPA using different search spaces. The first column represents the corresponding module we try to evaluate by fixing it with one OP in the reduced search space.}
	\label{tb-space-evaluation}
	\vspace{-10px}
\end{table}

\subsubsection{Spatial Aggregation Module}

To evaluate how the spatial aggregation module affects the performance,
we only search for the other three modules based on fixed aggregators RGCN and RGAT, 
which denoted as SPA-RGCN and SPA-RGAT, respectively. 
As shown in Table~\ref{tb-space-evaluation}, 
with fixed aggregators, 
SPA-RGCN and SPA-RGAT have a performance drop compared with SPA.
This indicatess that 
the diverse spatial aggregation modules can 
capture various topological information in different TKGs and significantly improve the model performance.

\subsubsection{Temporal Aggregation Module}

To evaluate the importance of searching for temporal aggregation module, 
we learn to design architectures with fixed temporal aggregation module instead.
In Table~\ref{tb-space-evaluation}, 
with the two predefined temporal aggregation operations, 
the degree of performance degradation is inconsistent across different datasets.
To be specific,
the performance drop is evident on SPA-IDENTITY for GDELT.
But for ICEWS14, 
the performance of SPA-GRU drops significantly compared to SPA.
This observation shows the importance of including temporal aggregation module in the search space.
Meanwhile, 
it shows that the temporal aggregation operations should also be data-specific for TKGC.

\subsubsection{Layer Connection and Layer Fusion Module}

In this section,
we evaluate the proposed Layer Connection and Layer Fusion Module, 
which are novel compared to existing GNN-based architectures for TKGC.
By fixing the skip-connection function as \texttt{LC\_SKIP}, 
we create the variant SPA-LC\_SKIP,
which means that we do not search for different skip-connection functions.
By fixing the layer fusion function as \texttt{LF\_SKIP}, 
we only preserve the output of last temporal aggregation module as entity representation.
This variant is denoted by SPA-LA\_SKIP, 
which means the outputs of intermediate layers are not used.

From Table~\ref{tb-space-evaluation},
we can see that

\begin{itemize}
	\item The performance drop of SPA-LC\_SKIP means that the spatial aggregation module can benefit from skip-connection, which have been shown in previous works~\cite{li2021deepgcns,li2020autograph}.
	\vspace{-5px}
	\item The performance drop of SPA-LA\_SKIP means that the outputs of intermediate layers are important for the final representation in temporal graph learning. Thus, it demonstrates the importance of the proposed Layer Fusion Module.
\end{itemize}

Taking all results in Table~\ref{tb-space-evaluation} into consideration, 
we can see that it is important for TKGC 
to search for combinations of operations 
from the four essential modules by SPA, 
which demonstrates the contribution of the proposed framework and the designed search space.

\subsection{Case Study}\label{sec:case-study}

We visualize the searched architectures 
on three benchmark datasets in Figure~\ref{fig-searched-archs}.
Especially,
the searched temporal aggregation modules contain more \texttt{IDENTITY} operations in ICEWS14,
while in GDELT more \texttt{SA} operations are searched.
This observation implies that 
for the ICEWS14 dataset, 
capturing complex temporal context may not be necessary 
in comparison to GDELT.

\begin{figure}[!h]
	\vspace{-10px}
	\subfloat{
		\begin{minipage}[t]{0.5\linewidth}
			\centering
			\includegraphics[width=0.95\linewidth]{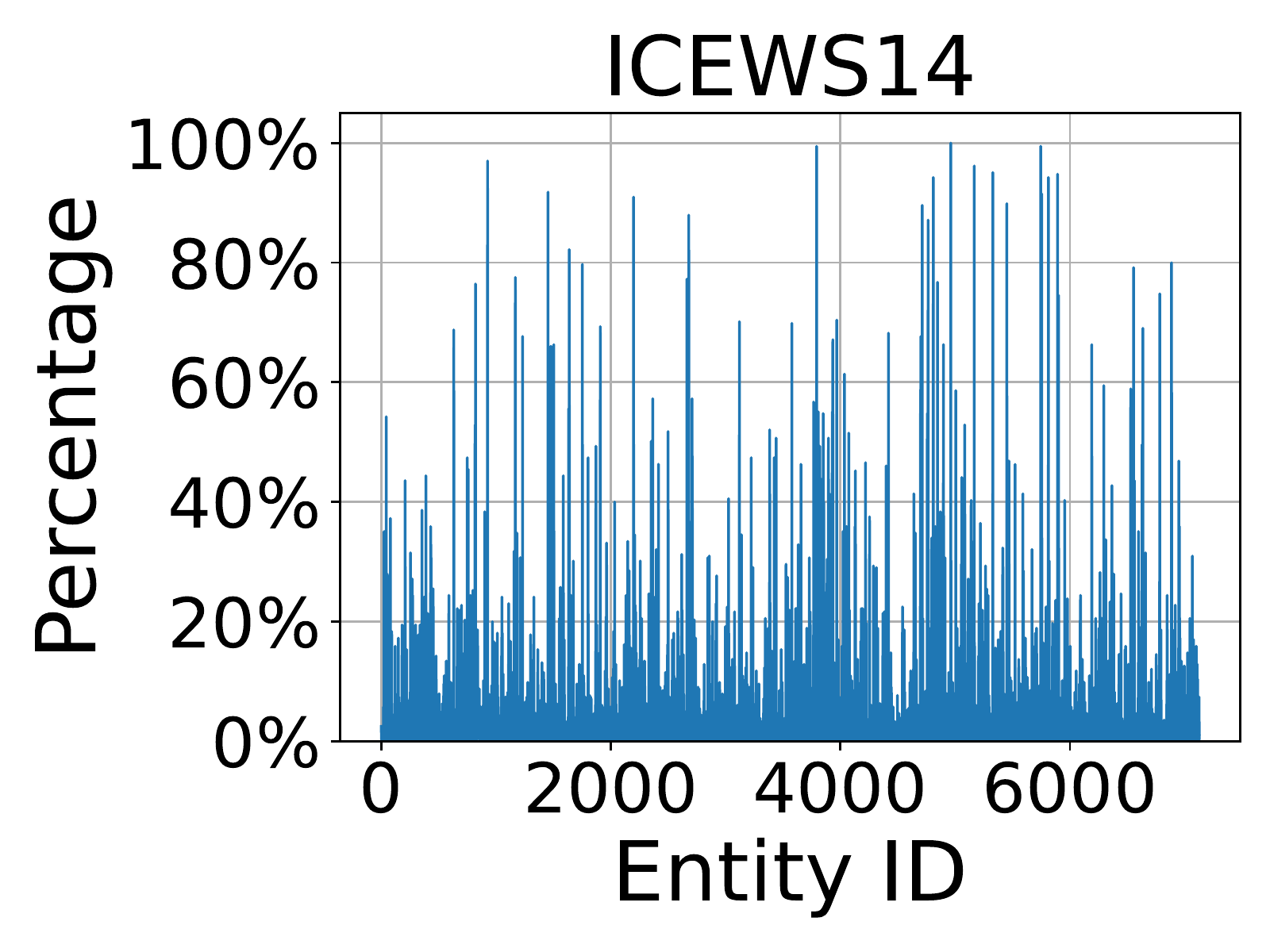}
		\end{minipage}%
	}%
	\subfloat{
		\begin{minipage}[t]{0.5\linewidth}
			\centering
			\includegraphics[width=0.95\linewidth]{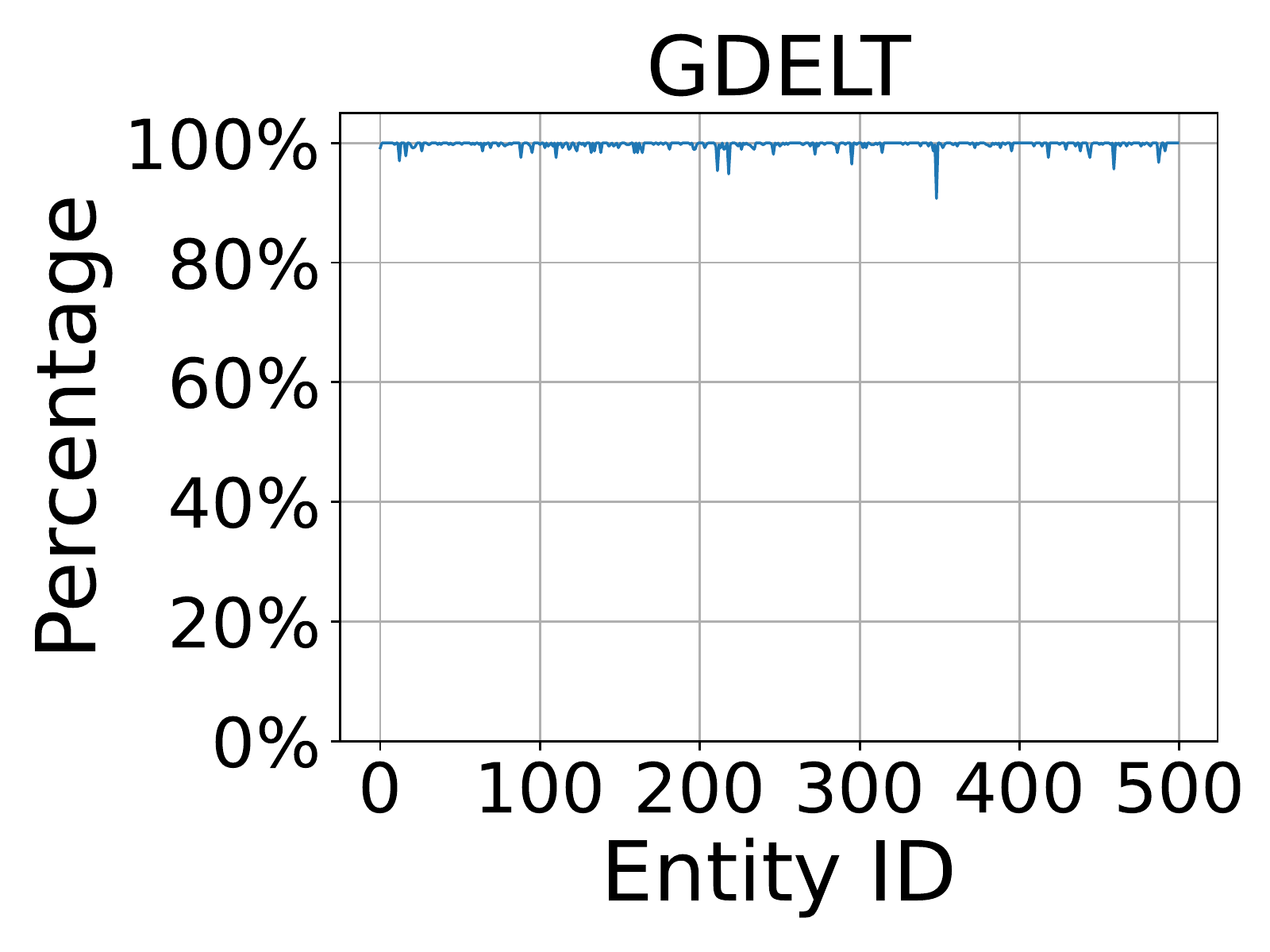}
		\end{minipage}%
	}
	\caption{Difference in temporal property between two datasets.
		The figure represents the proportion of timestamps when the entity is active\footnotemark to the total timestamps. As can be seen, entities in GDELT are much more active than those in ICEWS14.} 
	\label{fig-data_property}
	\vspace{-10px}
\end{figure}

To verify above conjecture, 
we compare the differences in temporal properties between ICEWS14 and GDELT.  
From the perspective of temporal properties, 
as shown in Figure~\ref{fig-data_property}, 
the activity frequency of entities on GDELT is much higher than ICEWS14.
This means that 
we do not need to design architectures 
with complicated sequential models 
for ICEWS14, 
but it is useful for GDELT.
This result confirms our conjecture and the importance of designing data-specific architectures for TKGC.

Taking into consideration these experimental results from Figure~\ref{fig-searched-archs} and \ref{fig-data_property},
it indicates the effectiveness of our method 
in finding data-specific architectures for TKGC. 

\footnotetext{An entity is active at a timestep if it has at least one neighboring entity in the same KG snapshot~\citep{wu-etal-2020-temp}.}

\section{Related Works}

\subsection{Temporal Knowledge Graph Completion (TKGC)}
 
In the literature, 
existing methods for TKGC can be roughly divided into two categories: the embedding-based method and the GNN-based method. 
Embedding-based methods~\citep{leblay2018deriving,dasgupta-etal-2018-hyte,goel2020diachronic,lacroix2020tensor, messner2022temporal}
design time-aware score functions to measure the correctness of quadruples in TKGs. 
Although embedding-based methods well capture the semantic patterns in TKGs, 
they fail to capture the more complex topological patterns.

Recently, 
with the success of graph neural networks (GNNs), 
GNN has achieved significant progress in temporal knowledge graph completion. 
TeMP~\citep{wu-etal-2020-temp},
uses structural encoder to obtain entity representations including multi-hop neighbor information 
and relies on temporal encoder to incorporate structural and temporal information into entity representation. 
T-GAP~\citep{jung2021learning} designs one temporal GNN to learn structural and temporal information on TKG, 
and another GNN to dynamically expand and prune the inference subgraph from the query entity $e_q$ by attention flow~\citep{xu2018modeling}. 
However, 
existing GNN-based methods use predefined structure and temporal encoder, 
which are difficult to adapt to various datasets. 

\subsection{Graph Neural Architecture Search}

Neural architecture search (NAS) aims to automatically find suitable neural architecture for the given dataset, 
which has been demonstrated as a promising technique in many research fields such as computer vision and neural language processing. 

More recently, 
some works focus on automatically designing GNNs by NAS. 
GraphNAS~\citep{gao2021graph}, AGNN~\citep{zhou2019auto} learn to design aggregation operation. 
AutoGraph~\citep{li2020autograph} learns to select the connections in each intermediate layer. 
SNAG~\citep{zhao2020simplifying} and SANE~\citep{huan2021search} search to select and fuse the features of intermediate layers in the output node. 
AutoGEL~\citep{wang2021autogel} focuses on designing intra-layer and inter-layer message passing GNN architectures automatically. 
However, 
no work applies NAS technique to design GNN for dynamic graphs or temporal knowledge graphs. 
To the best of our knowledge, 
SPA is the first method to learn data-specific GNN architectures for TKG completion.

\section{Conclusion}

In this paper, 
we propose a novel method SPA to automatically design data-specific architectures for TKGC task. We define a novel and expressive search space, 
in which different combinations of operations can capture various patterns of different TKGs. 
To enable efficient search on top of the search space, 
we adopt a flexible and effective search algorithm, 
which trains a simplified supernet in that each architecture is a single path, 
thus greatly reducing the GPU memory cost. 
To demonstrate the effectiveness of SPA for TKGC, 
we conduct extensive experiments on three datasets. 
The experimental results show that SPA can search SOTA data-specific architectures for TKGC.

For future work, 
we will explore more advanced NAS approaches to further improve the search efficiency of SPA. 
Besides, 
a promising direction is to explore how to efficiently search network architectures and hyper-parameters simultaneously.

\clearpage

\section*{Limitations}

There are two limitations for SPA. (1) SPA is focused on method design rather than system design. In the future, we will co-design the algorithm and the system to further improve the efficiency. (2) At present, SPA only search for data-specific architectures, while hyper-parameters are also important for TKGC. A promising direction is to explore how to efficiently search network architectures and hyper-parameters simultaneously.

\section*{Acknowledgements}
We thank the anonymous reviewers for their valuable comments. 
This work was supported in part by
the National Natural Science Foundation of China (No. U1803263),
the National Science Fund for Distinguished Young Scholarship of China (No. 62025602),  
Fok Ying-Tong Education Foundationm China (No. 171105), 
Key Technology Research and Development Program of Science and Technology-Scientific and Technological Innovation Team of Shaanxi Province (No. 2020TD-013),
and the XPLORER PRIZE. 

Q. Yao is sponsored by CCF-Baidu Open Fund and Tsinghua University-Foshan Institute of Advanced Manufacturing.

\bibliography{anthology,custom}
\bibliographystyle{acl_natbib}
\appendix

\section{Appendix}
\label{sec:appendix}

\subsection{Loss Function}
\label{sec:appendix-loss}

To train our TKGC model using score function, 
the model parameters are learned 
using gradient-based optimization in mini-batches.
Specifically.
for each quadruple $\eta = (s, r, o, t) \in \mathcal{D}^{+}$,
we sample a negative set of entities $\mathcal{D}^{-}_{\eta, o} = \{o' | (s, r, o', t) \not\in \mathcal{D}^{+}\}$.
Then,
we apply the cross-entropy loss function for object queries to train the model:
\begin{equation}\label{equ-loss_obj}
	\mathcal{L}_{\text{obj}} = - \sum_{(s,r,o,t) \in \mathcal{D}^{+}}\frac{\exp(\phi(s, r, o, t))}{\sum_{o' \in \mathcal{D}^{-}_{\eta, o}} \exp(\phi(s, r, o', t))}.
\end{equation}

Similarly, 
we can also obtain the loss for subject queries $\mathcal{L}_{\text{sub}}$. 
The final training loss is the sum of losses for two types of queries: $\mathcal{L} = \mathcal{L}_{\text{sub}} + \mathcal{L}_{\text{obj}}$.

\subsection{Dataset Statistics and Characteristics}
\label{sec:appendix-dataset}

The dataset statistics are summarized in Table~\ref{tab-dataset}.

\subsection{Implementation Details}
\label{sec:appendix-impl}

All the experiments are implemented in Python with the PyTorch framework~\citep{paszke2019pytorch} and run on a single NVIDIA RTX 3090 GPU with 24GB memory.

For Random, 
we use the Adam optimizer, 
set learning rate is 0.001, 
dropout rate = 0.1, and L2 norm to 0.0005.
We randomly sample 100 architectures from the designed search space and train them from scratch.   
After training finished, 
we select one candidate with the highest validation performance.

\begin{table*}[!h]
	\centering
	\begin{tabular}{cccccccc}
		\toprule
		\textbf{Dataset} &\textbf{\# entities} & \textbf{\# relations} &\textbf{\# time steps} &\multicolumn{1}{c}{\bm{$N_{train}$}}   & \multicolumn{1}{c}{\bm{$N_{valid}$}} & \multicolumn{1}{c}{\bm{$N_{test}$}} & \multicolumn{1}{c}{\bm{$N_{total}$}}\\
		\midrule
		ICEWS14 & 7,128 &230 &365 &72,826 &8,941 &8,963 &90,730 \\
		ICEWS05-15 & 10,488 & 251 & 4017 & 386,962 & 46,275 & 46,092 & 479,329 \\
		GDELT & 500 & 20 & 366 & 2,735,685 & 341,961 & 341,961 &3,419,607 \\
		\bottomrule
	\end{tabular}
	\caption{Statistics of ICEWS14, ICEWS05-15 and GDELT datasets. }
	\label{tab-dataset}
\end{table*}

\begin{table*}[!h]
	\centering 
	\setlength\tabcolsep{2pt}
	\begin{tabular}{cccccccc}
		\toprule
		\textbf{Dataset} & \textbf{Batch size} & \bm{$\tau$} & \makecell{\textbf{Head number of} \\\textbf{spatial module}} & \makecell{\textbf{Head number of} \\\textbf{temporal module}} &\makecell{\textbf{Embedding} \\ \textbf{size}} & \makecell{\textbf{Hidden} \\ \textbf{size}}  & \makecell{\textbf{Gradient}\\ \textbf{clipping}} \\
		\midrule
		ICEWS14 & 8 & 8 & 4 & 4 & 100 & 100 & 1\\
		\midrule
		ICEWS05-15 & 8 & 8 & 4 & 4 & 100 & 100 & 1\\
		\midrule
		GDELT & 2 & 4 & 4 & 4 & 100 & 100 & 1\\
		\bottomrule
	\end{tabular}
	\caption{Other hyperparameters setting for SPA during the search process.}
	\label{hyperparameters}
\end{table*}

For SPA, 
we set the epoch $T_1$ for supernet training is 800 and the epoch $T_2$ for architecture searching is 1000.
in each minibatch sample single path to train supernet. 
After training process is finished, 
we derive the candidate architecture with the highest validation performance from the supernet by random search. 
Repeat 5 times with different seeds, we can get 5 candidates.

Other hyperparameters settings for NAS methods during the search process are shown in Table~\ref{hyperparameters}.

The searched candidates are finetuned individually with the hyper-parameters shown in Table~\ref{tab-hyper-finetune}. 
In the stage of fine-tuning, 
we use the ReduceLROnPlateau scheduler.
Each method candidates 30 hyper steps. 
In each hyper step, a set of hyperparameters will be sampled from Table~\ref{tab-hyper-finetune} based on Hyperopt, and then generate final performance on the testing data. 

\begin{table}[hp]
	\centering
	\begin{tabular}{cc}
		\toprule
		\textbf{Hyperparameter} & \textbf{Value range}              \\ 
		\midrule
		\makecell{Head number \\of spatial module}  &  $\{2,4,8\}$                             \\ 
		\midrule
		\makecell{Head number \\of temporal module}  &        $\{2,4,8\}$                         \\ 
		\midrule
		Weight decay   &           $[10^{-5},10^{-3}]$         \\ 
		\bottomrule
	\end{tabular}
	\caption{Hyperparameters we used during the fine-tuning stage.}
	\label{tab-hyper-finetune}
\end{table}

\end{document}